\documentclass[conference]{IEEEtran}
\IEEEoverridecommandlockouts

\usepackage{cite}
\usepackage{amsmath,amssymb,amsfonts}
\usepackage{algorithmic}
\usepackage{graphicx}
\usepackage{textcomp}
\usepackage{xcolor}
\usepackage{multirow}
\usepackage{tcolorbox}
\usepackage{booktabs}
\usepackage[numbers]{natbib}
\usepackage[hidelinks]{hyperref}

\def\BibTeX{{\rm B\kern-.05em{\sc i\kern-.025em b}\kern-.08em
    T\kern-.1667em\lower.7ex\hbox{E}\kern-.125emX}}

\begin{document}

\title{UBIWEAR: An end-to-end, data-driven framework for intelligent physical activity prediction to empower mHealth interventions}

\author{\IEEEauthorblockN{Asterios Bampakis\IEEEauthorrefmark{1},
Sofia Yfantidou\IEEEauthorrefmark{2} and Athena Vakali\IEEEauthorrefmark{3}}
\IEEEauthorblockA{School of Informatics,
Aristotle University of Thessaloniki, Greece\\
Email: \IEEEauthorrefmark{1}maasterio@csd.auth.gr,
\IEEEauthorrefmark{2}syfantid@csd.auth.gr,
\IEEEauthorrefmark{3}avakali@csd.auth.gr}}

\IEEEpubid{\begin{minipage}{\textwidth} Copyright \copyright 2022 IEEE International Conference \\ 
on E-health Networking, Application \& Services (HealthCom), \\
\url{https://doi.org/10.1109/HealthCom54947.2022.9982730}
\end{minipage}}

\maketitle

\IEEEpubidadjcol

\begin{abstract}
It is indisputable that physical activity is vital for an individual's health and wellness. However, a global prevalence of physical inactivity has induced significant personal and socioeconomic implications. In recent years, a significant amount of work has showcased the capabilities of self-tracking technology to create positive health behavior change. This work is motivated by the potential of personalized and adaptive goal-setting techniques in encouraging physical activity via self-tracking. To this end, we propose UBIWEAR, an end-to-end framework for intelligent physical activity prediction, with the ultimate goal to empower data-driven goal-setting interventions. To achieve this, we experiment with numerous machine learning and deep learning paradigms as a robust benchmark for physical activity prediction tasks. To train our models, we utilize, ``MyHeart Counts", an open, large-scale dataset collected in-the-wild from thousands of users. We also propose a prescriptive framework for self-tracking aggregated data preprocessing, to facilitate data wrangling of real-world, noisy data. Our best model achieves a MAE of 1087 steps, 65\% lower than the state of the art in terms of absolute error, proving the feasibility of the physical activity prediction task, and paving the way for future research.
\end{abstract}


\section{Introduction}
According to the World Health Organization (WHO), physical inactivity is one of the leading risk factors for noncommunicable diseases and death worldwide, inducing substantial personal and societal cost \cite{world2019global}. On a personal level, it significantly increases the risk of cancer, heart disease, and diabetes, and it is estimated that up to five million deaths per year could be prevented if the global population were sufficiently active. Deteriorating population health also comes with a growing societal burden in terms of medical care and loss of productivity. Recent estimates show that physical inactivity has led to US\$ 54 billion expense for the health system and US\$ 14 billion in indirect economic losses in the US alone, while worldwide, 1–3\% of national health care expenditures can be attributed to physical inactivity \cite{world2019global}. 

A significant challenge towards reducing the global prevalence of physical inactivity is encouraging individuals who are not sufficiently active to alter their behavior and include physical activity in their daily routine. To this end, research has shown that interactive technology can be strategically designed to motivate desirable behavior change, such as regular exercise and healthy nutrition \cite{gouveia2015we}, for better population health and wellness. 
At the same time, investing in technological tools and resources for promoting regular physical activity can directly contribute to many of the United Nations' 2030 Sustainable Development Goals \cite{nations2015transforming}. Overall, across all settings, there are opportunities for digital mHealth innovations to harness the potential of data to promote, support, monitor, and sustain health behavior change, focusing on physical activity. 

The U.S. Department of Health and Human Services recommends the equivalent of at least 150 minutes of moderate-intensity aerobic activity each week to rip the benefits of regular exercise \cite{pa_guidelines}. Simplifying these guidelines into something concrete and relatable, such as daily step counts, is an easy way for the majority of the population to understand and achieve them. Specifically, daily step counts between 7,000 and 9,000 steps can result in health benefits similar to achieving the recommended amounts of moderate-to-vigorous exercise \cite{kraus2019daily}. To this end, considerable research efforts have been made for developing effective technological interventions to help people achieve the recommended step counts. Amongst the most common and successful persuasive design techniques utilized in such interventions is goal-setting, i.e., setting a daily number of target steps for the user to achieve \cite{yfantidou2021self}. 

The most straightforward goal-setting approach, used in most commercial physical activity trackers, is the \textit{fixed goal approach}, where the system sets a fixed goal for the user. However, this approach may lead to unrealistic goals, as it does not consider the singularity of an individual's behavior. At the same time, research shows that a \textit{personalized and adaptive goal approach} performs better in increasing adherence and physical activity levels \cite{adaptive_personalized_advisor}, by tailoring the system to enhance motivational appeal. In this approach, the system monitors the daily behavior of the user, such as physical activity, sleep, and stress levels, and then personalizes their step goal based on a combination of factors. At the same time, it adapts over time to incorporate possible changes in user behavior.

Previous works have attempted to tackle the task of personalized and adaptive goal-setting by employing statistical models for time-series forecasting or domain-expertise-based estimates of step goals tailored to different user groups (see Section~\ref{related-work}). However, the focus is currently shifted towards more intelligent approaches, which provide state-of-the-Art (SotA) performance for the task of physical activity prediction \cite{adaptive_personalized_advisor,dimitrios_vasdekis_2022_5839727}. Nevertheless, the field of physical activity prediction for adaptive goal-setting is still in its infancy, and SotA approaches suffer from various limitations discussed below:

\textbf{Small-scale \& Subjective Experimental Data (\textit{L1}):} Prior work relies on small-scale datasets that are constructed and designed according to conceptual or theoretical interests with articulated research questions, and collected by users participating in ongoing research, which can lead to a distorted reflection of the actual levels of their daily activity \cite{zhou_fukuoka_mintz_goldberg_kaminsky_flowers_aswani_2018};

\textbf{Challenging Data Wrangling (\textit{L2}):} 
There are only scattered guidelines concerning preprocessing techniques suitable for handling the idiosyncrasies of real-world, noisy time-series data generated by personal informatics self-tracking systems, contrary to popular domains, such as natural language processing or computer vision;

\textbf{Lack of Physical Activity Prediction Benchmarking (\textit{L3}):} Limited studies exist in the area of predicting physical activity levels by exploiting machine learning and deep learning approaches for enabling personalized and adaptive goal-setting. Even then, the exact architectures and hyperparameters utilized remain unspecified, and a comprehensive benchmarking of relevant approaches is yet to be published;

\textbf{Reproducibility \& Reusability Issues (\textit{L4}):} Due to the limitations introduced by closed-source data and code repositories, reproducibility of published results and reusability of existing code and models for the task of personalized goal-setting is close to infeasible at this moment. 

\begin{figure}[htb!]
    \vspace{-3mm}
    \centering
    \includegraphics[width=.55\linewidth]{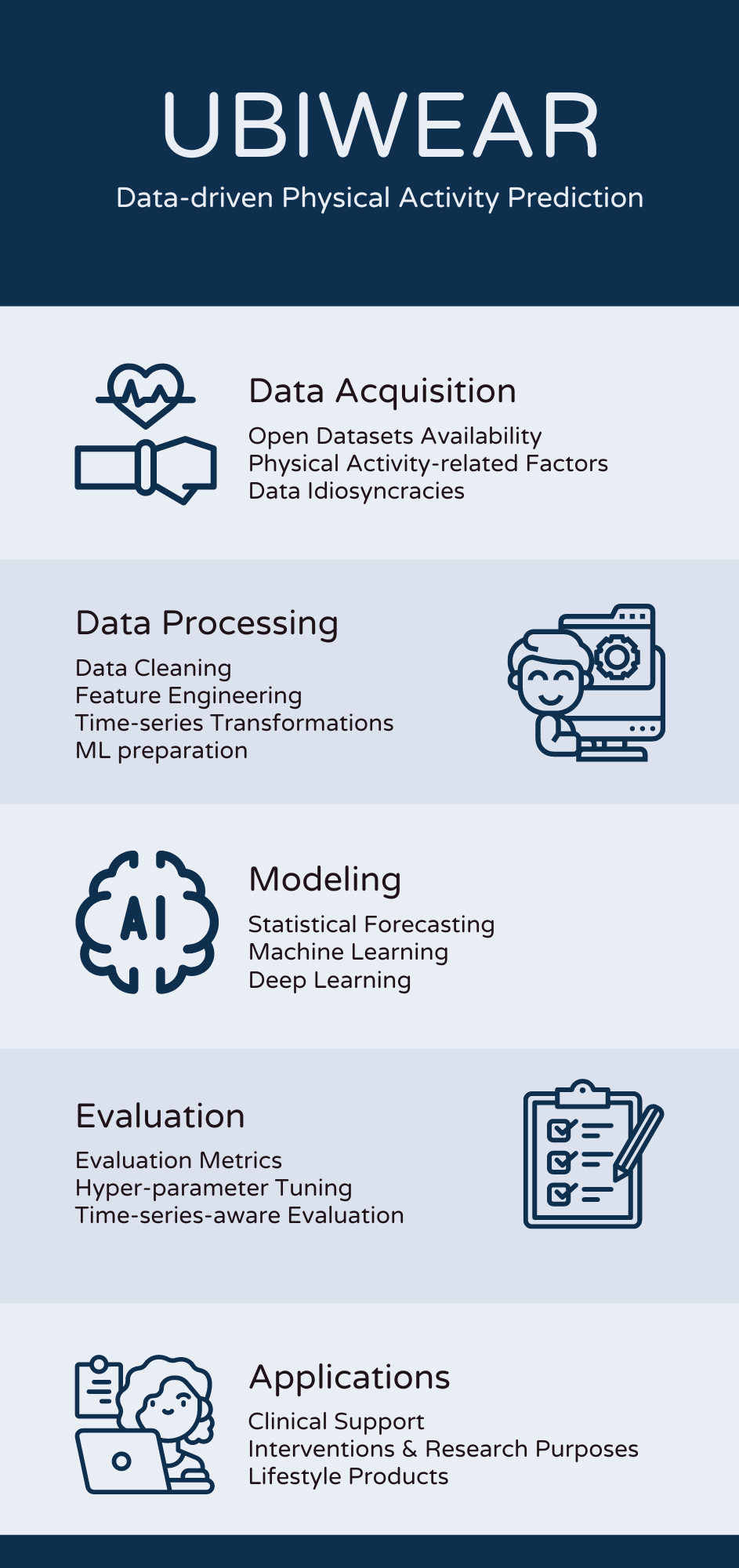}
    \caption{The UBIWEAR framework covers the path of physical activity data from its acquisition to its application scenarios.}
    \label{fig:ubiwear}
\end{figure}

Motivated by the issues above, this work proposes UBIWEAR, a five-step, data-driven framework for the task of physical activity prediction with the end goal of enabling personalized and adaptive goal-setting interventions via self-tracking. The framework comprises the complete process from physical activity data acquisition to end-user applications. Figure~\ref{fig:ubiwear} depicts UBIWEAR's pipeline. Our contributions are structured around the framework's steps:

\textbf{\textit{C1} - In-the-wild Data Exploitation:} We discuss the availability of open datasets for physical activity prediction, and the idiosyncrasies of self-tracking data. Ultimately, we utilize the openly accessible ``MyHeart Counts'' dataset \cite{hershman2019physical}, containing real-world, large-scale physical activity patterns for thousands of users, allowing us to better capture different segments of the population and build generalizable prediction models. The dataset has been collected in-the-wild, presenting an objective view of the users' behavior (L1).

\textbf{\textit{C2} - Self-tracking Data Processing Guidelines:} We introduce a set of prescriptive guidelines on how to process aggregated data from self-tracking devices, accompanied by a Python library release
\footnote{https://pypi.org/project/ubiwear/}. Derived from our exhaustive experimentation, we propose specific adaptation methodologies for traditional preprocessing techniques explicitly designed to handle self-tracking data idiosyncrasies. Our goal is for these guidelines to provide a more standardized definition of processing self-tracking data (L2).

\textbf{\textit{C3} - Physical Activity Prediction Benchmarking \& Evaluation:} We experiment with six different learning paradigms for physical activity prediction, from machine learning to advanced deep learning architectures, and benchmark their performance for this complex learning task (L3). Through the experimentation with more advanced architectures, UBIWEAR achieves a MAE (Mean absolute error) of 1087 steps, 65\% lower in terms of absolute error than that of the SotA model \cite{mohammadi2019neural}, proving the feasibility of intelligent physical activity prediction.

\textbf{\textit{C4} - Open Access Data and Code:} We purposefully work with an open-access dataset and publicly share our codebase
\footnote{https://github.com/stergiosbamp/deep-physical-activity-prediction} to enable the reproducibility of our results. To further facilitate future interdisciplinary research, we also adopt containerization and offer step-by-step guides on how to download, store and preprocess the data, as well as on how to reproduce our results with the uploaded pre-trained models (L4). 

We structure the remaining of this paper as follows: Section~\ref{related-work} discusses the related literature in physical activity prediction and personalized goal-setting. Sections \ref{meth1}-\ref{meth5} introduce the UBIWEAR framework from data acquisition to self-tracking data processing, modeling and evaluation to applications of insights. Finally, Section~\ref{conclusions} concludes the paper and delineates ideas for future work.

\label{introduction}

\section{Related Work: Intelligent Physical Activity Prediction for Personalized Goal-setting}
An emerging body of work is trying to tackle the lack of intelligent goal-setting systems by publishing studies of statistical, rule-based, machine learning- or deep learning-based physical activity prediction solutions that take into account historic data to provide realistic and achievable exercise goals.

\citet{van2018enhancing} developed a context-aware coaching system that adapts to each individual by taking into consideration factors such as context, and historical patterns of behavior. Similarly, ``Active2Gether" \cite{klein2017active2gether} is a coaching system that utilizes the behavioral characteristics of users and social comparison to provide personalized feedback. However, the most fundamental component of both systems, namely the reasoning engine that determines what type of support the user should receive, relies on a purely rule-based approach rather than a predictive one, limiting its extensibility potential. With regards to statistical forecasting approaches, \citet{zhou2018evaluating} investigated the effectiveness of personalized automated adaptive step goals, in comparison to fixed 10000 daily step goals. Their findings revealed that the intervention group had statistically significant more daily steps than the fixed steps goal group, but only few details were provided about the behavioral algorithm used.

Closer to our work, \citet{dijkhuis2018personalized} developed a machine learning system that aimed to predict whether or not an individual will meet their daily step goal. The authors benchmarked eight traditional machine learning models, but oversimplified the task of predicting activity goals as a binary decision. To bridge this gap, \citet{dimitrios_vasdekis_2022_5839727} developed a framework for machine learning-based physical activity prediction formulated as a regression task. While they explored a variety of significant contextual factors, their small-scale dataset might not be representative of the general population, while it prohibited them from bechmarking data-hungry SotA deep learning approaches. \citet{mohammadi2019neural} experimented both with traditional machine learning and a neural network architecture to predict a dynamically adjusted daily number of steps based on personal, environmental, and social factors. However, their experimentation was solely based on a small-scale dataset and they did not provide any information concerning the deep learning architectures and hyperparameters used, limiting the reproducibility and reusability of their approach. Their SotA ``BRIDGE'' model achieved a MAE of 1672 steps, 65\% higher than our proposed approach.

It is evident that only a limited number of studies exploit machine learning or deep learning techniques for the task of physical activity prediction for enabling personalized and adaptive goal-setting, even though SotA models provide significantly superior performance in similar tasks, such as human activity recognition or sleep stage classification. This gap motivated our work and guided UBIWEAR's methodology as will be presented in detail in the following sections (\ref{meth1}-\ref{meth5}).
\label{related-work}

\section{Data Acquisition\label{meth1}}
A motivation serving as the backbone of our framework is the peculiarities that materialize the problem of predicting future physical activity with machine learning. Although it can fall under the abstract category of time-series forecasting problems, the self-tracking data idiosyncrasies perplex the problem and thus require particular attention. The UBIWEAR framework is designed with the following in mind:
\begin{itemize}
    \item Raw data collected from sensors or wearables are sampled in arbitrary frequencies contrary to traditional time-series data.
    \item Physical activity datasets are guaranteed to be a multi-subject time-series sequences.
    \item Multiple input devices and overlapping records due to different users can introduce duplicates.
    \item Consecutive records cannot be assumed in a real-world scenario as users do not necessarily track their activities on a daily basis (no-wear time).
\end{itemize}

On top of the above-mentioned peculiarities, data acquisition for physical activity prediction is challenging; collecting proprietary data requires a significant monetary and time investment, while open datasets are limited and include diverse data modalities. Popular small-scale (in terms of sample size) datasets include the Extrasensory dataset (N=60 users) containing a wide range of sensor data combined with activity and location labels \cite{vaizman2017recognizing}; and the StudentLife dataset (N=48) comprised of continuous sensor data, self-reports, and various pre-post surveys combined with activity, mental health and academic performance labels \cite{wang2014studentlife}. On a larger-scale, the FitRec datasets contain user sport records from Endomondo, including multi-modal sequential sensor data and contextual information \cite{ni2019modeling}. In this study, we utilize a recent, large-scale, in-the-wild dataset from the ``MyHeart Counts Cardiovascular Health Study'' \cite{hershman2019physical}. In this study, participants contributed health data via an iPhone application, related to physical activity, fitness, sleep, and overall health for studying the patterns and any potential associations between them. We keep a subset of recordings relevant to our task, i.e., the number of steps that were recorded by the device at a particular time. Our final dataset contains 9,154,490 examples for total of 4,747 unique users of diverse gender, ethnicity, physical and health condition. 

\section{Data Processing for Self-tracking Data\label{meth2}}
\begin{figure}[htb!]
    \vspace{-3mm}
    \centering
    \includegraphics[width=.65\linewidth, keepaspectratio]{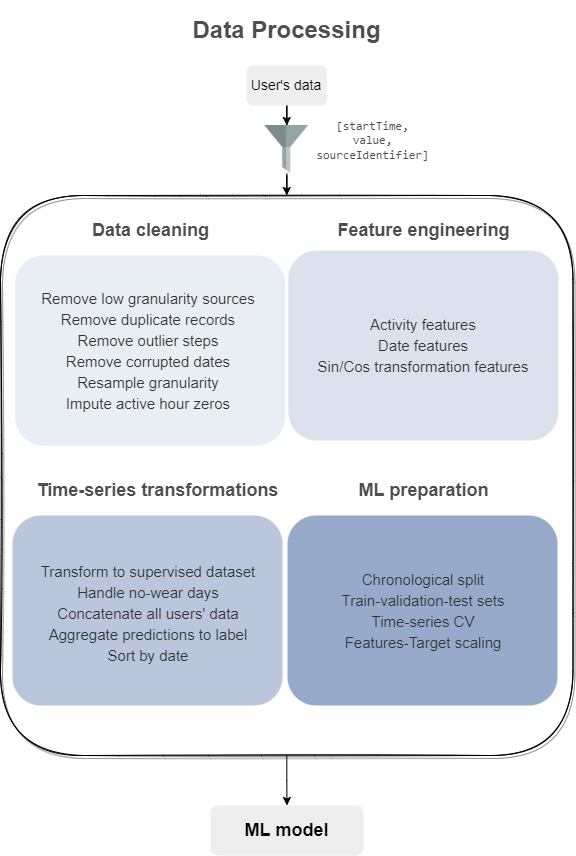}
    \caption{An illustration of the data processing pipeline.}
    \label{fig:data-preprocessing}
    \vspace{-2mm}
\end{figure}

As discussed in Section~\ref{introduction}, there are no concrete guidelines for the preprocessing of aggregated self-tracking data, despite the idiosyncrasies of the task. Following our data acquisition step, we move to data processing by introducing a series of prescriptive guidelines as part of the UBIWEAR framework (Figure \ref{fig:data-preprocessing}). They consist of four components, namely Data Cleaning, Feature Engineering, Time-series Transformations, and Machine Learning Preparation. Each generic component is composed of traditional data pre-processing steps, \textit{tweaked for the self-tracking data domain}. 

\noindent{\textbf{Data Cleaning.}} We initiate data cleaning by \textbf{removing duplicate records}, i.e., records with the same timestamp for a single user. Given the multi-subject and multi-device nature of self-tracking data, duplicate elimination significantly differs from traditional time-series processing. Specifically, self-tracking duplicates removal should be performed on a user-by-user basis, while caution should be given in cases of multiple input devices, as overlapping records can introduce duplicates. We proceed by \textbf{removing the outlier dates and values}. Outlier dates can arise from missing or corrupted dates in the raw data. To overcome this issue, we preserve the dates that correspond to the intervention duration, and drop any remaining missing dates. With regards to outlier step values, there are different sub-cases of outliers, and the decision of how to define them must be chosen carefully. In this work, we adopt the quantile method, for the removal of outlier values and avoid aggressive cut-offs. Also, time-series analysis requires that our time-series data be recorded in fixed-time intervals, so we need to \textbf{achieve a unified granularity}. Self-tracking resampling should be performed only after verifying common granularity for all input sources. In case of arbitrary sampling frequencies, record drops or value splits should precede resampling. The first approach (record drops), adopted in this work, is to drop low frequency records since they represent a small percentage of our dataset. A more sophisticated approach (value split) can be to divide these coarser granularity values by the number of daily active hours, and assign it to the hourly records per example. Next, during \textbf{user filtering}, we filter out any users that do not have the necessary number of days for the requested window size. Note that the larger the requested window size, the stricter the filtering criteria, and hence less users are included in the final dataset. Finally, we proceed with \textbf{zero values imputation}. Zero values observations can be reasonable in the night hours, but in the daytime inactivity is usually caused by forgotten self-tracking devices. Based on that intuition, zero values imputation should be performed only for active hours.\\
\noindent{\textbf{Feature Engineering.}} Feature Engineering includes the transformation of raw data into meaningful variables. A common practice in a majority of time-series problems is to enhance the feature space with features that are derived from timestamps (e.g., hour, day, week, month), as well as their \textbf{\textit{sin} and \textit{cos} transformations} to capture the cyclic nature of time \cite{martucci1994symmetric}. Furthermore, features such as bank holiday and or weekend can facilitate time-series forecasting \cite{taylor2018forecasting}. \\
\noindent{\textbf{Time-series Transformations.}} The \textbf{sliding window technique} allows treating the problem of physical activity prediction as a time-series forecasting problem. Specifically, self-tracking windowing should consider sliding windows for daily granularity and tumbling windows for hourly granularity to avoid excessive overlap of records. Care should be given to the target variable to represent daily (or higher) aggregate counts, in the case of hourly (or finer) granularity.\\
\noindent{\textbf{Machine Learning Preparation.}} The Machine Learning preparation step allows us to prepare our data to be fed into our learning models, including a \textbf{train-validation-test split}. However, in self-tracking data we need to respect chronological order i.e. no shuffling. Since we utilize neural networks, \textbf{feature and target scaling} is almost a mandatory step to ensure that the gradient descent moves smoothly towards the minima and that the steps for gradient descent are updated at the same rate for all features \cite{sola1997importance}. 

\section{Modeling for Physical Activity Prediction\label{meth3}}
An important contribution of our work is the exhaustive benchmarking of the various learning models in the previously unexplored domain of physical activity prediction for enabling personalized and adaptive goal-setting. 
Under this spectrum, we explore both machine learning and deep learning approaches to provide a comprehensive comparison. 

In the machine learning category, we study three families of algorithms: linear models, tree models, and ensemble models. For the case of linear models, we choose the Ridge regressor \cite{mcdonald2009ridge} to minimize the multicollinearity of our data. Due to the hourly granularity, the number of features that arise is high enough. Thus, we believe that the regularization penalty to the loss function would be beneficial. In the case of tree models, we opt for decision trees, where we select the squared error function to measure the quality of a split which minimizes the L2 loss using the mean of each terminal node.
The third family of algorithms we explore is ensemble learning, and specifically the Gradient Boosting Regression \cite{friedman2002stochastic}, because of its capability to reduce both bias and variance in the data. 

To achieve an end-to-end benchmarking of physical activity prediction, we also investigate how deep learning models of various architectures fit this task. We scrutinize and apply four different types of architectures. Multilayer Perceptrons (MLPs), 1-D Convolutional Neural Networks (CNNs) and Recurrent Neural Networks (RNNs) with long short-term memory (LSTM) cells. MLPs is a classic category of neural networks that can be applied in time-series problems by having one output node in the last layer which emits the actual regression value. 1-D CNNs are considered state-of-the-art in the signal processing domain, with their remarkable ability to automatically extract locality features in the time dimension. Last but not least, RNNs are well-suited for approaching time-series problems by catching temporal dependencies along with the input sequence. With regards to the deep learning models, we designed a high-level overview of the optimal architectures that yielded the best performance, following an exhaustive architectures analysis and hyperparameter tuning via trial and error discussed in Section \ref{exp2}.
\begin{figure*}[]
\minipage{0.25\textwidth}
     \includegraphics[width=.8\linewidth]{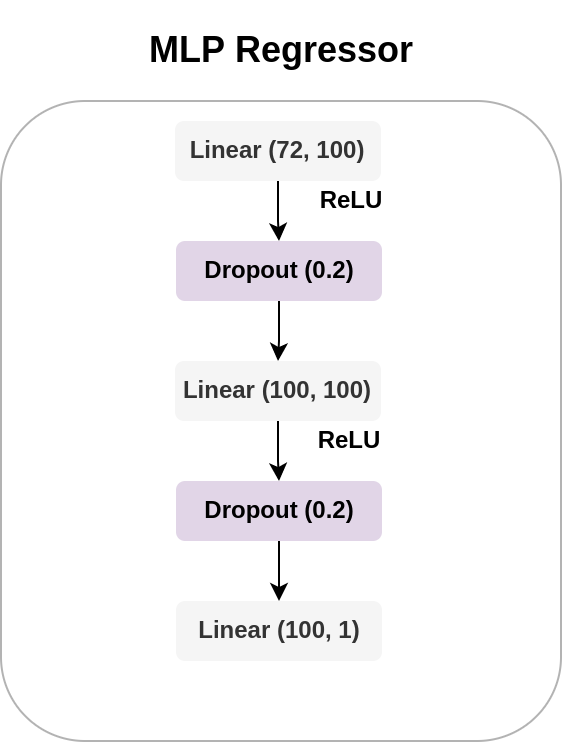}
\endminipage\hfill
\minipage{0.25\textwidth}
  \includegraphics[width=.8\linewidth]{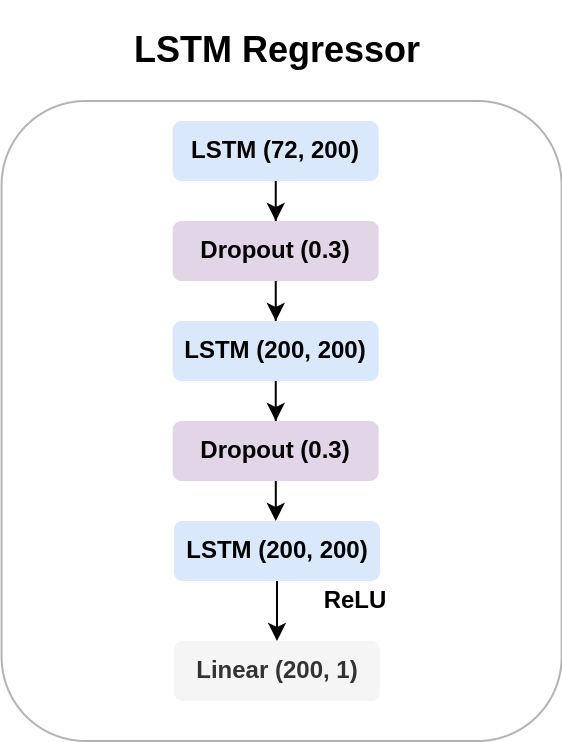}
\endminipage\hfill
\minipage{0.25\textwidth}%
  \includegraphics[width=.8\linewidth]{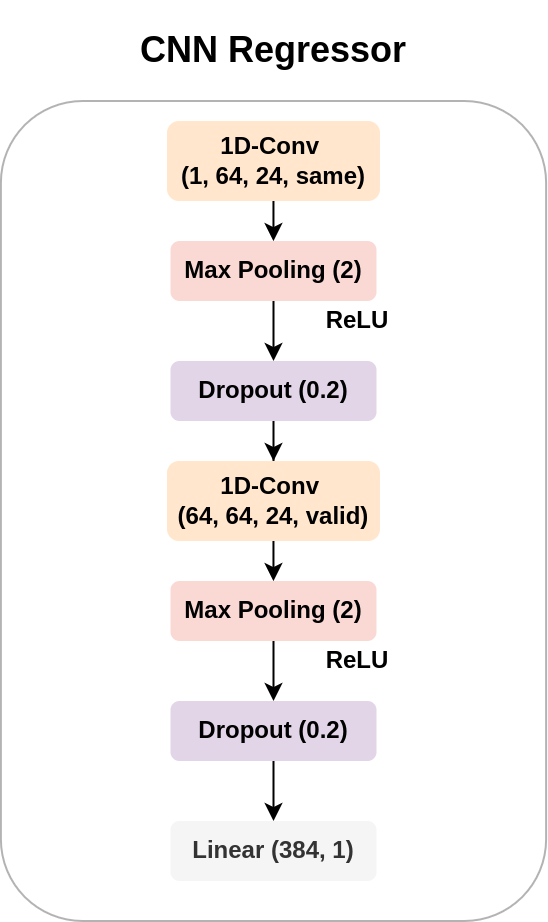}
\endminipage
 \caption{The final architectures of the MLP (left), RNN/LSTM (center), and 1D-CNN (right) regressors, respectively.\label{fig:mlp-arch}}
\end{figure*}

\section{Evaluation \& Implications\label{meth4}}
In this section we present our experimentation with data processing (\ref{exp1}) and modeling benchmarking (\ref{exp2}).

\subsection{Effects of Data Processing on Physical Activity Prediction \label{exp1}}
To conclude on the final dataset, we conduct a number of experiments regarding UBIWEAR's processing steps. Our data processing evaluation is relied on a time-aware train-validation-test split of our data, and we select the Ridge regressor as the baseline model in this experiment due to its ability to provide quick and accurate results.

With regards to \textbf{optimal granularity \& window size}, we experiment with hourly and daily aggregation of the data, while we simultaneously experiment with the window size (1-6 days) to obtain more fine-grained results (12 combinations in total). Given that the goal is to provide a model that can be deployed in the real-world, waiting to gather more than a week's data is neither user-friendly nor practical. The difference in error for every window size (MAE$>$500 steps) allows us to deduce that hourly granularity outperforms daily granularity in every case. In terms of window size, the performance of a 3-day window is comparable to this of larger windows. The improvement in MAE and MdAE between the 2-day and 3-day windows (MAE$_2=1417$ and MAE$_3=1361$) is higher than the improvement between the 3-day and larger windows (MAE$_4=1327$, MAE$_5=1300$ and MAE$_6=1281$). This evidence in conjointment to the practicality of smaller windows drives us to select three as the optimal window size.

In addition, we investigate whether the addition of \textbf{date features and cyclical transformations} can benefit the models compared to step features only. We have observed that in terms of MdAE the steps only dataset yields the best results (MdAE$_{steps\_only}=392.14$, MdAE$_{steps\_date\_features}=446.73$, MdAE$_{steps\_cyclic\_features}=574.41$), while the MAE scores are comparable. We assume that this is due to the high overlap of different users' records in our dataset. In other words, by extending the feature space with the date features for the same timestamps but different user behavior, we essentially introduce noise to our machine learning models by diverting attention from the actual activity data. 

We finally experiment with \textbf{removing the outlier values} based on the quantiles method with $q=0.05$. Based on our results, we can deduce that indeed outliers removal boosts the performance of the model, by excluding individual abnormal patterns from our data and provides more conclusive predictions in unknown cases (MAE$_{keep\_outliers}=1887.16$ and MAE$_{remove\_outliers}=1307.01$).

Overall, we proceed to the models' benchmarking with a dataset consisting of 3-day windows of hourly granularity, without imputation of missing data, and having undergone outlier removal (both for dates and values with a $q=0.05$), as this combination of preprocessing steps shows the best performance for the ``MyHeart Counts'' dataset. 

\subsection{Benchmarking Deep Learning: Does it match the hype?\label{exp2}}
As discussed previously, for the modeling procedure we divide our dataset into three sets: training, validation and test set (no shuffling). We utilize the validation set for the hyper-parameter tuning of the models to provide the optimal results.

\begin{table}[htb!]
    \caption{Performance benchmarking for all ML \& DL models.\label{tab:ml-dl-benchmarking}}
    \centering
    \begin{tabular}{@{}llll@{}}
    \toprule
    \textbf{Model} & \textbf{Dataset} & \multicolumn{2}{c}{\textbf{Metrics}} \\ \midrule
     &  & \textbf{MAE} & \textbf{MdAE} \\ \midrule
    \multirow{3}{*}{Ridge} & Train & 1372.889 & 458.038 \\
     & Test & 1359.766 & 390.271 \\ \midrule
    \multirow{3}{*}{Decision Tree} & Train & 1428.906 & 592.652 \\
     & Test & 1478.086 & 224.101 \\ \midrule
    \multirow{3}{*}{Gradient Boosting} & Train & 1209.139 & 463.878 \\
     & Test & 1222.738 & 135.668 \\ \midrule
    \multirow{3}{*}{MLP} & Train & 1148.293 & 296.582 \\
     & Test & 1094.822 & \textbf{0.082} \\ \midrule
    \multirow{3}{*}{CNN} & Train & 1167.130 & 284.548 \\
     & Test & 1099.581 & 11.110 \\ \midrule
    \multirow{3}{*}{RNN (LSTM)} & Train & 1171.620 & 280.274 \\
     & Test & \textbf{1087.838} & 0.855 \\ \bottomrule
    \end{tabular}
\end{table}
For the classic machine learning models, we find the best hyperparameters per paradigm through a grid search, time-series-based cross-validation technique with 5 splits. For the deep learning models, we identify the optimal hyperparameters by experimenting with various combinations of learning rate, dropout, hidden units, batch size, optimizer, kernel size, pooling and stride, per each architecture. Figure \ref{fig:mlp-arch} delineates the final architectures for the MLP, RNN and CNN, accordingly.

Table \ref{tab:ml-dl-benchmarking} shows the overall modeling outcomes against the train and test datasets. The best performing model is found to be the RNN achieving 1087 MAE and only 1 step of MdAE, confirming the assumption that RNNs are able to capture temporal dependencies in time-series problems. Equally well performance is delivered by a 3-layered feed-forward network (MLP) with almost similar results on the hold-out set. The lower MdAE from MLP can be justified on the intuition that predictions from this architecture are more robust to outlier examples, in contrast to the ones from RNN architecture. However, this difference is negligible since RNN can predict the overall cases more efficiently.
The CNN proved unsuitable for extracting features from the convolution operation although we utilized a fairly large kernel size given the temporal features. Despite that, the deep learning models deliver almost 1.25 times better performance than a simple regression model (e.g., Ridge) in terms of MAE, but the required computational resources and training time are significantly higher, a fact that should be considered before deducing that neural networks are the proper fit for every task. This is also more apparent in the case of deployment of such models for delivering interventions based on predictive modeling in wearable devices, where that computational resources are limited.

\section{UBIWEAR Applications\label{meth5}}
A vital question emerging is how our work and ultimately machine learning-based solutions for physical activity prediction can be embraced in real-world scenarios. Through our research, we provide some indicative use cases in which UBIWEAR can be applied.

First and foremost, by predicting personalized physical activity levels from the unique users' data, we can infer adaptive and challenging future step goals by increasing the predicted step goal by 10\%. Research has shown that activity goals proximate to personal levels for each individual can be advantageous for encouraging them and maintaining motivation to complete daily physical activity goals \cite{adaptive_personalized_advisor}. Even industrial wearables software and mHealth services for exercise can integrate our pre-trained models and inject into their system our predictions, to adapt to the specificities of each person and deliver intelligent future step goals. Furthermore, in a similar manner UBIWEAR can be adopted in randomized controlled trials (RCTs) to study behavioral changes in different groups of users. The outcome of such a procedure can benefit behavioral transformation interventions in a majority of healthcare applications that promote healthy lifestyle habits from different disciplines such as medicine or pharmacy. Therefore they can focus on the development of the clinical support systems and alleviate the cost of conducting an end-to-end, time-consuming pipeline and designing a machine learning system from scratch for that purpose.

\label{methodology-experimentation}

\section{Discussion \& Conclusions\label{conclusions}}
Despite our contributions, it is essential to acknowledge some limitations of our work. Given the little related work and the lack of open-source benchmarking datasets in the domain, it is infeasible to provide a systematic comparison with relevant algorithms. We hope that by focusing on reproducibility, we pave the way for future research in the field. Additionally, one should be cautious about applying the pre-trained models to different populations. The ``MyHeart Counts'' dataset, despite its great potential, has certain limitations in terms of demographic diversity. Hence, the applicability of the trained models for diverse subgroups of the population needs to be assessed before real-world deployment.

In summary, this paper addresses one of the most unexplored domains in wearable analytics through the introduction of UBIWEAR, a data-driven framework for physical activity prediction for personalized and adaptive goal-setting. Under this spectrum, we utilize real-world in-the-wild data originating from the ``MyHeart Counts" study (C1). The data exploited are at least thirty times larger than other related studies, which allows us to better capture different user patterns and provide more robust and generalized prediction algorithms. We explore both machine and deep learning approaches to provide end-to-end benchmarking. We propose a 3-stacked RNN (LSTM) as a SotA model for this task, achieving 1,087 MAE and 1 MdAE error in step counts given only three previous days' data as input  (C3). Our approach establishes a high standard and prepares the ground for future work. The most time-consuming procedure was the pre-processing pipeline tailored to self-tracking data peculiarities. This led to the design of a set of prescriptive guidelines and the release of an open-sourced Python library to help researchers and practitioners with the highly demanding task of data processing of ubiquitous self-tracking data (C2). Last but not least, all of our work is publicly available for data collection, data ingestion, pre-processing, experimentation, model training, and evaluation with pre-trained models uploaded to facilitate the reproducibility of this study (C4).
\label{discussion-conclusions}

\section*{Acknowledgment}
This project has received funding from the European Union’s Horizon 2020 research and innovation programme under the Marie Skłodowska-Curie grant agreement No 813162. The content of this paper reflects only the authors' view and the Agency and the Commission are not responsible for any use that may be made of the information it contains.

\bibliographystyle{IEEEtranN}
\bibliography{IEEEabrv, bibliography}

\end{document}